\documentclass{article}

\usepackage{microtype}
\usepackage{graphicx}
\usepackage{subfigure}
\usepackage{booktabs} 
\usepackage{Definitions}
\usepackage{enumitem}
\usepackage{diagbox}
\usepackage{multirow}
\usepackage[subtle]{savetrees}
\usepackage{hyperref}

\graphicspath{{./figs/}}

\usepackage[accepted]{icml2018}

\icmltitlerunning{Adversarial Attack on Graph Structured Data}

\begin{document}

\twocolumn[
\icmltitle{Adversarial Attack on Graph Structured Data}



\icmlsetsymbol{equal}{*}

\begin{icmlauthorlist}
\icmlauthor{Hanjun Dai}{gatech}
\icmlauthor{Hui Li}{ant}
\icmlauthor{Tian Tian}{th}
\icmlauthor{Xin Huang}{ant}
\icmlauthor{Lin Wang}{ant}
\icmlauthor{Jun Zhu}{th}
\icmlauthor{Le Song}{gatech,ant}
\end{icmlauthorlist}

\icmlaffiliation{gatech}{Georgia Institute of Technology}
\icmlaffiliation{ant}{Ant Financial}
\icmlaffiliation{th}{Tsinghua University}

\icmlcorrespondingauthor{Hanjun Dai}{hanjundai@gatech.edu}

\icmlkeywords{Machine Learning, ICML}

\vskip 0.3in
]



\printAffiliationsAndNotice{}  

\begin{abstract}
Deep learning on graph structures has shown exciting results in various applications. However, few attentions have been paid to the robustness of such models, in contrast to numerous research work for image or text adversarial attack and defense. In this paper, we focus on the adversarial attacks that fool the model by modifying the combinatorial structure of data. We first propose a reinforcement learning based attack method that learns the generalizable attack policy, while only requiring prediction labels from the target classifier. Also, variants of genetic algorithms and gradient methods are presented in the scenario where prediction confidence or gradients are available. 
We use both synthetic and real-world data to show that, a family of Graph Neural Network models are vulnerable to these attacks, in both graph-level and node-level classification tasks. We also show such attacks can be used to diagnose the learned classifiers. 
\end{abstract}

\section{Introduction}

\begin{figure*}[t]
\centering
\includegraphics[width=1.0\textwidth]{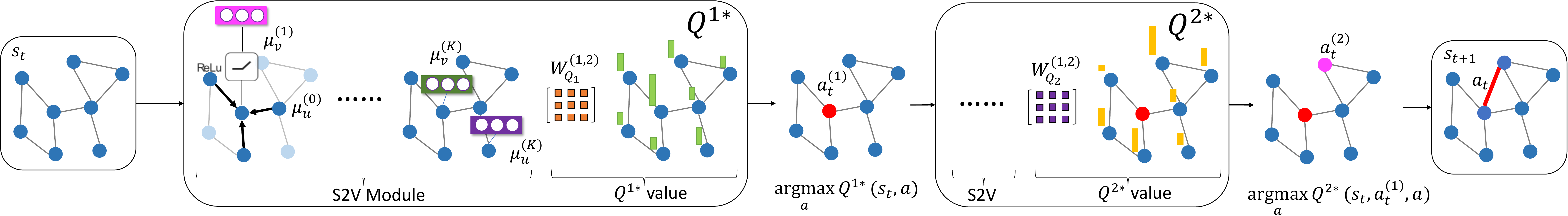}
\vspace{-4mm}
\caption{Illustration of applying hierarchical Q-function to propose adversarial attack solutions. Here adding a single edge $a_t$ is decomposed into two decision steps $a_t^{(1)}$ and $a_t^{(2)}$, with two Q-functions $Q^{1*}$ and $Q^{2*}$, respectively. \label{fig:dqn}}
\end{figure*}

Graph structure plays an important role in many real-world applications. 
Representation learning on the structured data with deep learning methods has shown promising results in various applications, including drug screening~\citep{DuvMacIpaBometal15}, protein analysis~\citep{HamYinLes17}, knowledge graph completion~\citep{TriDaiWanSon17}, \etc. 

Despite the success of deep graph networks, the lack of interpretability and robustness of these models make it risky for some financial or security related applications. As analyzed in~\citet{AkoTonKou15}, the graph information is proven to be important in the area of risk management. A graph sensitive evaluation model will typically take the user-user relationship into consideration: a user who connects with many high-credit users may also have high credit. Such heuristics learned by the deep graph methods would often yield good predictions, but could also put the model in a risk. 
A criminal could try to disguise himself by connecting other people using Facebook or Linkedin. Such `attack' to the credit prediction model is quite cheap, but the consequence could be severe. Due to the large number of transactions happening every day, even if only one-millionth of the transactions are fraudulent, fraudsters can still obtain a huge benefit. However, few attentions have been put on domains involving graph structures, despite the recent advances in adversarial attacks and defenses for other domains like images~\citep{GooShlSze14} and text~\citep{JiaLiang17}. 

So in this paper, we focus on the graph adversarial attack for a set of graph neural network(GNN)~\citep{ScaGorTsoHagetal09} models. These are a family of supervised~\citep{DaiDaiSon16} models that have achieved state-of-the-art results in many transductive tasks~\citep{KipWel16} and inductive tasks~\citep{HamYinLes17}.
Through experiments in both node classification and graph classification problems, we will show that the adversarial samples do exist for such models. And the GNN models can be quite vulnerable to such attacks. 

However, effectively attacking graph structures is a non-trivial problem.
Different from images where the data is continuous, the graphs are discrete. Also the combinatorial nature of the graph structures makes it much more difficult than text. Inspired by the recent advances in combinatorial optimization~\citep{BelPhaLeNoretal16, DaiKhaZhaDiletal17}, we propose a reinforcement learning based attack method that learns to modify the graph structure with only the prediction feedback from the target classifier. The modification is done by sequentially add or drop edges from the graph. A hierarchical method is also used to decompose the quadratic action space, in order to make the training feasible. Figure~\ref{fig:dqn} illustrates this approach. We show that such learned agent can also propose adversarial attacks for new instances without access to the classifier. 

Several different adversarial attack settings are considered in our paper. When more information from the target classifier is accessible, a variant of the gradient based method and a genetic algorithm based method are also presented. Here we mainly focus on the following three settings:
\begin{itemize}[leftmargin=*,nosep,nolistsep]
\item \textbf{white box attack (WBA)}: in this case, the attacker is allowed to access any information of the target classifier, including the prediction, gradient information, \etc. 
\item \textbf{practical black box attack (PBA)}: in this case, only the prediction of the target classifier is available. When the prediction confidence is accessible, we denote this setting as PBA-C; if only the discrete prediction label is allowed, we denote the setting as PBA-D.
\item \textbf{restrict black box attack (RBA)}:  this setting is one step further than PBA. In this case, we can only do black-box queries on some of the samples, and the attacker is asked to create adversarial modifications to other samples. 
\end{itemize}

As we can see, regarding the amount of information the attacker can obtain from the target classifier, we can sort the above settings as WBA $>$ PBA-C $>$ PBA-D $>$ RBA. For simplicity, we focus on the non-targeted attack, though it is easy to extend to the targeted attack scenario. 

In Sec~\ref{sec:background}, we first present the background about GNNs and two supervised learning tasks. Then in Sec~\ref{sec:graph_attack} we formally define the graph adversarial attack problem. Sec~\ref{sec:rl_attack} presents the attack method \textit{RL-S2V} that learns the generalizable attack policy over the graph structure. We also propose other attack methods with different levels of access to the target classifier in Sec~\ref{sec:other_attack}. We experimentally show the vulnerability of GNN models in Sec~\ref{sec:experiment}, and also present a way of doing defense against such attacks.  

\section{Background}
\label{sec:background}
A set of graphs is denoted by $\Gcal = \{G_i\}_{i=1}^{N}$, where $|\Gcal| = N$. Each graph $G_i = (V_i, E_i)$ is represented by the set of nodes $V_i = \{v^{(i)}_j\}_{j=1}^{|V_i|}$ and edges $E_i = \{\eb^{(i)}_j\}_{j=1}^{|E_i|}$. Here the tuple $\eb^{(i)}_j = (\eb^{(i)}_{j, 1}, \eb^{(i)}_{j, 2}) \in V_i \times V_i$ represents the edge between node $\eb^{(i)}_{j, 1}$ and $\eb^{(i)}_{j, 2}$. 
In this paper, we focus on undirected graphs, but it is straightforward to extend to directed ones. Optionally, the nodes or edges can have associated features. We denote them as $x(v^{(i)}_j) \in \RR^{D_{node}}$ and $w(\eb^{(i)}_j) = w(\eb^{(i)}_{j,1}, \eb^{(i)}_{j, 2})\in \RR^{D_{edge}}$, respectively. 

This paper works on attacking the graph supervised classification algorithms. Here two different supervised learning settings are considered:

\noindent \textbf{Inductive Graph Classification:} We associate each graph $G_i$ with a label $y_i \in \Ycal = \{1, 2, \ldots, Y\}$, where $Y$ is the number of categories. The dataset $\Dcal^{(ind)} = \{(G_i, y_i)\}_{i=1}^N$ is represented by pairs of graph instances and graph labels. This setting is \textit{inductive} since the test instances will never be seen during training. Examples of such task including classifying the drug molecule graphs according to their functionality. In this case, the classifier $f^{(ind)} \in \Fcal^{(ind)}: \Gcal \mapsto \Ycal $ is optimized to minimize the following loss:
\begin{equation}
	\Lcal^{(ind)} = \frac{1}{N} \sum_{i=1}^N L(f^{(ind)}(G_i), y_i)
	\label{eq:ind_loss}
\end{equation}
 where $L(\cdot, \cdot)$ is the cross entropy by default.  

\noindent \textbf{Transductive Node Classification:} In node classification setting, a target node $c_i \in V_i$ of graph $G_i$ is associated with a corresponding node label $y_i \in \Ycal$. The classification is on the nodes, instead of the entire graph. We here focus on the transductive setting, where only a single graph $G_0 = (V_0, E_0)$ is considered in the entire dataset. That is to say, $G_i = G_0, \forall G_i \in \Gcal$. It is transductive since test nodes (but not their labels) are also observed during training. Examples in this case include classifying papers in a citation database like Citeseer, or entities in a social network like Facebook. Here the dataset is represented as $\Dcal^{(tra)} = \{(G_0, c_i, y_i)\}_{i=1}^N$, and the classifier $f^{(tra)}(\cdot ; G_0) \in \Fcal^{(tra)}: V_0 \mapsto \Ycal$ minimizes the following loss:
\begin{equation}
	\Lcal^{(tra)} = \frac{1}{N} \sum_{i=1}^N L(f^{(tra)}(c_i; G_0), y_i)
	\label{eq:tra_loss}
\end{equation}

When not causing confusion, we will overload the notations $\Dcal = \{(G_i, c_i, y_i)\}_{i=1}^N$ to represent the dataset, and $f \in \Fcal$ in either settings. In this case, $c_i$ is implicitly omitted in inductive graph classification setting; While in transductive node classification setting, $G_i$ always refers to $G_0$ implicitly. 

\noindent \textbf{GNN family models}

The Graph Neural Networks (GNNs) define a general architecture for neural network on graph $G = (V, E)$. This architecture obtains the vector representation of nodes through an iterative process: 
\begin{eqnarray}
	\mu_{v}^{(k)} = & h^{(k)}\big(\{w(u, v), x(u), \mu_{u}^{(k-1)}\}_{u \in \Ncal(v)}, \nonumber \\
& x(v), \mu_{v}^{(k-1)}\big), k \in \{1, 2, \ldots, K\}
	\label{eq:gnn}
\end{eqnarray}
where $\Ncal(v)$ specifies the neighborhood of node $v \in V$. The initial node embedding $\mu_v^{(0)} \in \RR^{d}$ is set to zero. For simplicity, we denote the outcome node embedding as $\mu_v = \mu_v^{(K)}$. To obtain the graph-level embedding from node embeddings, a global pooling is applied over the node embeddings. 

The vanilla GNN model runs the above iteration until convergence. But recently, people find a fixed number of propagation steps $T$ with various different parameterizations~\citep{LiTarBroZem15, DaiDaiSon16, GilSchRilVineetal17, LeiJinRegJaa17} work quite well in various applications. 

\section{Graph adversarial attack}
\label{sec:graph_attack}

Given a learned classifier $f$ and an instance from the dataset $(G, c, y) \in \Dcal$, the graph adversarial attacker $g(\cdot, \cdot): \Gcal \times \Dcal \mapsto \Gcal$ asks to modify the graph $G = (V, E)$ into $\tilde{G} = (\tilde{V}, \tilde{E})$, such that
\begin{eqnarray}
	\max_{\tilde{G}} && \II( f(\tilde{G}, c) \neq y) \nonumber \\
	s.t. &&  \tilde{G} = g(f, (G, c, y)) \nonumber \\
	&& \Ical(G, \tilde{G}, c) = 1 .
	\label{eq:attack_obj}
\end{eqnarray}
Here $\Ical(\cdot, \cdot, \cdot): \Gcal \times \Gcal \times V \mapsto \{0, 1\}$ is an equivalency indicator that tells whether two graphs $G$ and $\tilde{G}$ are equivalent under the classification semantics. 

In this paper, we focus on the modifications to the discrete structures. 
The attacker $g$ is allowed to add or delete edges from $G$ to construct the new graph. Such type of actions are rich enough, since adding or deleting nodes can be performed by a series of modifications to the edges. Also modifying the edges is harder than modifying the nodes, since choosing a node only requires $O(|V|)$ complexity, while naively choosing an edge requires $O(|V|^2)$. 

Since the attacker is aimed at fooling the classifier $f$, instead of actually changing the true label of the instance, the equivalency indicator should be defined first to restrict the modifications an attacker can perform. We use two ways to define the equivalency indicator:

\begin{itemize}[leftmargin=*,nosep,nolistsep]
	\item[1)] Explicit semantics. In this case, a gold standard classifier $f^*$ is assumed to be accessible. Thus the equivalency indicator $\Ical(\cdot, \cdot, \cdot)$ is defined as:
	\begin{equation}
		\label{eq:r_explicit}
		\Ical(G, \tilde{G}, c) = \II( f^*(G, c) = f^*(\tilde{G}, c) ),
	\end{equation}
	where $\II(\cdot) \in \{0, 1\}$ is an indicator function. 
	\item[2)] Small modifications. In many cases when explicit semantics is unknown, we will ask the attacker to make as few modifications as possible within a neighborhood graph:
	\begin{align}
		\label{eq:r_few}
		\Ical(G, \tilde{G}, c) = & \II( |(E - \tilde{E}) \cup (\tilde{E} - E) | < m ) \nonumber \\
		& \cdot \II( \tilde{E} \subseteq \Ncal(G, b)) ).
	\end{align} 
	In the above equation, $m$ is the maximum number of edges that allowed to modify, and $\Ncal(G, b) = \{(u, v): u, v \in V, d^{(G)}(u, v) <= b \}$ defines the $b$-hop neighborhood graph, where $d^{(G)}(u, v) \in \{1, 2, \ldots\}$ is the distance between two nodes in graph $G$. 
	\end{itemize}
	
	Take an example in friendship networks, a suspicious behavior would be adding or deleting many friends in a short period, or creating the friendship with someone who doesn't share any common friend. The ``small modification'' constraint eliminates the possibility of above two possibilities, so as to regulate the behavior of $g$. 
	With either of the two realizations of robust classifier $r$, it is easy to enforce the attacker. Each time when an invalid modification proposed, the classifier can simply ignore such move.

	Below we first introduce our main algorithm, \textit{RL-S2V}, for learning attacker $g$ in Section~\ref{sec:rl_attack}. Then in Section~\ref{sec:other_attack}, we present other possible attack methods under different scenarios. 
	
\subsection{Attacking as hierarchical reinforcement learning}
\label{sec:rl_attack}

Given an instance $(G, c, y)$ and a target classifier $f$, we model the attack procedure as a Finite Horizon Markov Decision Process $\Mcal^{(m)}(f, G, c, y)$. The definition of such MDP is as follows: 

\begin{itemize}[leftmargin=*,nosep,nolistsep]
\item \textbf{Action} As we mentioned in Sec~\ref{sec:graph_attack}, the attacker is allowed to add or delete edges in the graph. So a single action at time step $t$ is $a_t \in \Acal \subseteq V \times V$. However, simply performing actions in $O(|V|^2)$ space is too expensive. We will shortly show how to use hierarchical action to decompose this action space. 
\item \textbf{State} The state $s_t$ at time $t$ is represented by the tuple $(\hat{G}_t, c)$, where $\hat{G_t}$ is a partially modified graph with some of the edges added/deleted from $G$. 
\item \textbf{Reward} The purpose of the attacker is to fool the target classifier. So the non-zero reward is only received at the end of the MDP, with reward being
\begin{equation}
	r\big((\tilde{G}, c)\big) = \begin{cases}
		1:  f(\tilde{G}, c) \neq y \\ 
		-1: f(\tilde{G}, c) = y \\
	\end{cases}
\end{equation}
In the intermediate steps of modification, no reward will be received. That is to say, $r(s_t, a_t) = 0, \forall t = 1, 2, \ldots, m-1$. In PBA-C setting where the prediction confidence of the target classifier is accessible, we can also use $r\big((\tilde{G}, c)\big) = \Lcal(f(\tilde{G}, c), y)$ as the reward.
\item \textbf{Terminal} 	Once the agent modifies $m$ edges, the process stops. For simplicity, we focus on the MDP with fixed length. In the case when fewer modification is enough, we can simply let the agent to modify the dummy edges. 
\end{itemize}

Given the above settings, a sample trajectory from this MDP will be: $(s_1, a_1, r_1, \ldots, s_m, a_m, r_m, s_{m+1})$, where $s_1 = (G, c)$, $s_t = (\hat{G}_t, c), \forall t \in \{2, \ldots, m\}$ and $s_{m+1} = (\tilde{G}, c)$. The last step will have reward $r_m = r(s_m, a_m) = r\big((\tilde{G}, c)\big)$ and all other intermediate rewards are zero: $r_t = 0, \forall t \in \{1, 2, \ldots, m-1\}$. 
Since this is a discrete optimization problem with a finite horizon, we use Q-learning to learn the MDPs. In our preliminary experiments we also tried with policy optimization methods like Advantage Actor Critic, but found Q-learning works more stable. So below we focus on the modeling with Q-learning. 

Q-learning is an off-policy optimization where it fits the Bellman optimality equation directly as below: 
\begin{equation}
	Q^*(s_t, a_t) = r(s_t, a_t) + \gamma \max_{a'} Q^*(s_{t+1}, a').
	\label{eq:bellman_orig}
\end{equation}
This implicitly suggests a greedy policy:
\begin{equation}
	\pi(a_t | s_t; Q^*) = \arg \max_{a_t} Q^*(s_t, a_t).
	\label{eq:greedy_policy}
\end{equation}
In our finite horizon case, $\gamma$ is fixed to 1. Note that directly operating the actions in $O(|V|^2)$ space is too expensive for large graphs. Thus we propose to decompose the action $a_t \in V \times V$ into $a_t = (a_t^{(1)}, a_t^{(2)})$, where $a_t^{(1)}, a_t^{(2)} \in V$. Thus a single edge action $a_t$ is decomposed into two ends of this edge. The hierarchical Q-function is then modeled as below:
\begin{eqnarray}
	Q^{1*}(s_t, a_t^{(1)}) = & \max_{a_t^{(2)}} Q^{2*}(s_t, a_t^{(1)}, a_t^{(2)})  \nonumber \\
	Q^{2*}(s_t, a_t^{(1)}, a_t^{(2)}) = & r\big(s_t, a_t = (a_t^{(1)}, a_t^{(2)})\big) + \nonumber \\ 
	& \max_{a_{t+1}^{(1)}} Q^{1*}(s_t, a_{t+1}^{(1)}) .
	\label{eq:bellman_two}
\end{eqnarray}
In the above formulation, $Q^{1*}$ and $Q^{2*}$ are two functions that implement the original $Q^*$. An action is considered as completed only when a pair of $(a_t^{(1)}, a_t^{(2)})$ is chosen. Thus the reward will only be valid after $a_t^{(2)}$ is made. It is easy to see that such decomposition has the same optimality structure as in Eq~(\ref{eq:bellman_orig}), but making an action would only require $O(2 \times |V|) = O(|V|)$ complexity. Figure~\ref{fig:dqn} illustrates this process. 

Take a further look at Eq~(\ref{eq:bellman_two}), since only the reward in last time step is non-zero, and also the budget of modification $m$ is given, we can explicitly unroll the Bellman equations as:
\begin{eqnarray}
	Q_{1,1}^*(s_1, a_1^{(1)}) = & \max_{a_1^{(2)}} Q_{1,2}^*(s_1, a_1^{(1)}, a_1^{(2)})  \nonumber \\
	Q_{1,2}^*(s_1, a_1^{(1)}, a_1^{(2)}) = & \max_{a_{2}^{(1)}} Q_{2,1}^*(s_2, a_{2}^{(1)}) \nonumber \\
	\ldots \nonumber \\ 
	Q_{m,1}^*(s_m, a_m^{(1)}) = & \max_{a_m^{(2)}} Q_{m, 2}^*(s_m, a_m^{(1)}, a_m^{(2)})  \nonumber \\
	Q_{m,2}^*(s_m, a_m^{(1)}, a_m^{(2)}) = & r( \tilde{G}, c) 
	\label{eq:bellman_unrol}
\end{eqnarray}

To make notations compact, we still use $Q^* = \{Q_{t, 1 | 2}^*\}_{t=1}^m$ to denote the Q-function. Since each sample in the dataset defines an MDP, it is possible to learn a separate Q function for each MDP $M_{i}^{(m)}(f, G_i, c_i, y_i), i = 1, \ldots, N$. However, we here focus on a more practical and challenging setting, where only one $Q^*$ is learned. The learned Q-function is thus asked to generalize or transfer over all the MDPs: 
\begin{equation}
	\max_{\thetab} \sum_{i=1}^N \EE_{t, a =\arg\max_{a_t} Q^*(a_t | s_t; \thetab)} [r\big((\tilde{G}_i, c_i)\big)],
\end{equation}
where $Q^*$ is parameterized by $\thetab$. Below we present the parameterization for such $Q^*$ that generalizes over MDPs. 

\subsubsection{Parameterization of $Q^{*}$}

From above, we can see the most flexible parameterization would be implementing $2 \times m$ time-dependent Q functions. However, we found two distinct parametrization is typically enough, \ie, $Q_{t,1}^* = Q^{1*}, Q_{t,2}^* = Q^{2*}, \forall t$. 

Since the $Q$ function is scoring the nodes in the state graph, it is natural to use GNN family models for parameterization, in order to learn a generalizable attacker. Specifically, $Q^{1*}$ is parameterized as:
\begin{equation}
	Q^{1*}(s_t, a_t^{(1)}) = W_{Q_1}^{(1)} \sigma \big(W_{Q_1}^{(2)\top} [\mu_{a_t^{(1)}}, \mu(s_t)] \big),
\end{equation}
where $\mu_{a_t^{(1)}}$ is the embedding of node $a_t^{(1)}$ in graph $\hat{G}_t$, obtained by structure2vec (S2V)~\citep{DaiDaiSon16}:
\begin{equation}
	\mu_v^{(k)} = \text{relu}(W_{Q_1}^{(3)}x(v) + W_{Q_1}^{(4)}\sum_{u \in \Ncal(v)} \mu_u^{(k-1)} ),
\end{equation}
where $\mu_v = \mu_v^{(K)}$ and $\mu_v^{(0)} = 0$. Also $\mu(s_t) = \mu(\hat{G}_t = (\hat{V}_t, \hat{E}_t), c)$ is the representation of entire state tuple:
 \begin{equation}
	\mu(s_t) =  \begin{cases}
		\sum_{v \in \hat{V}} \mu_{v} :  \text{graph attack} \\ 
		[\sum_{v \in \Ncal_{\hat{G}_t}(c, b)} \mu_{v}, \mu_{c}]: \text{node attack}\\
	\end{cases}
\end{equation}
In node attack scenario, the state embedding is taken from the $b$-hop neighborhood of node $c$, denoted as $\Ncal_{\hat{G}_t}(c, b)$. The parameter set of $Q^{1*}$ is $\theta_1 = \{W_{Q_1}^{(i)}\}_{i=1}^4$. $Q^{2*}$ is parameterized similarly with parameter $\theta_2$, with an extra consideration of the chosen node $a_t^{(1)}$:
\begin{equation}
	Q^{2*}(s_t, a_t^{(1)}, a_t^{(2)}) = W_{Q_2}^{(1)} \sigma \big(W_{Q_2}^{(2)\top} [\mu_{a_t^{(1)}}, \mu_{a_t^{(2)}}, \mu(s_t)] \big)
\end{equation}
We denote this method as \textit{RL-S2V} since it learns a Q-function parameterized by S2V to perform attack. 

\subsection{Other attacking methods}
\label{sec:other_attack}

The \textit{RL-S2V} is suitable for black-box attack and transfer. However, for different attack scenarios, other algorithms might be preferred. We first introduce \textit{RandSampling} that requires least information in Sec~\ref{sec:rand_sampling}; Then in Sec~\ref{sec:grad_attack}, a white-box attack \textit{GradArgmax} is proposed; Finally the \textit{GeneticAlg}, which is a kind of evolutionary computing, is proposed in Sec~\ref{sec:genetic_alg}.

\subsubsection{Random sampling}
\label{sec:rand_sampling}

This is the simplest attack method that randomly adds or deletes edges from graph $G$. When an edge modification action $a_t = (u, v)$ is sampled, we will only accept it when it satisfies the semantic constraint $\Ical(\cdot, \cdot, \cdot)$. It requires the least information for attack. Despite its simplicity, sometimes it can get good attack rate. 

\subsubsection{Gradient based white box attack}
\label{sec:grad_attack}

\begin{figure}[t]
\centering
\includegraphics[width=0.45\textwidth]{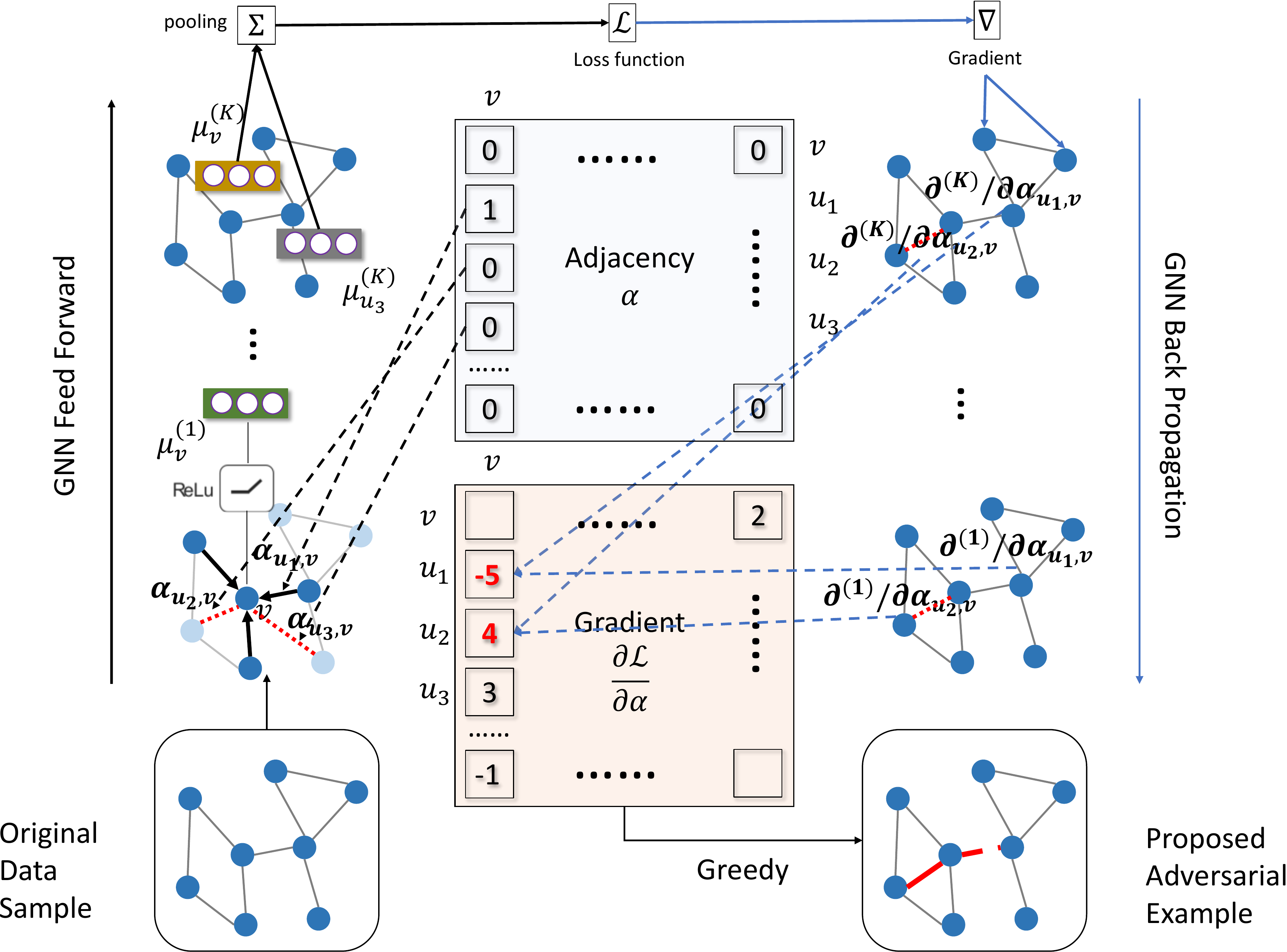}
\vspace{-4mm}
\caption{Illustration of graph structure gradient attack. This white-box attack adds/deletes the edges with maximum gradient (with respect to $\alpha$) magnitudes. \label{fig:grad_attack}}
\end{figure}

Gradients have been successfully used for modifying continuous inputs, \eg, images. However, taking gradient with respect to a discrete structure is non-trivial. Recall the general iterative embedding process defined in Eq~(\ref{eq:gnn}), we associate a coefficient $\alpha_{u, v}$ for each pair of $(u, v) \in V \times V$: 
\begin{eqnarray}
	\mu_{v}^{(k)} = h^{(k)} \big(& \{\alpha_{u, v} \big[w(u, v), x(u), \mu_{u}^{(k-1)}\big]\}_{u \in \Ncal(v)} \cup \nonumber \\
	& \{\alpha_{u', v} \big[w(u', v), x(u'), \mu_{u'}^{(k-1)}\big]\}_{u' \notin \Ncal(v)}, \nonumber \\
& x(v), \mu_{v}^{(k-1)}\big), k \in \{1, 2, \ldots, K\}
	\label{eq:gnn_alpha}
\end{eqnarray}
Let $\alpha_{u, v} = \II(u \in \Ncal(v))$. That is to say, $\alpha$ itself is the binary adjacency matrix. It is easy to see that the above formulation has the same effect as in Eq~(\ref{eq:gnn}). However, such additional coefficients give us the gradient information with respect to each edge (either existing or non-existing):
\begin{equation}
	\frac{\partial \Lcal}{\partial \alpha_{u, v}} = \sum_{k=1}^K \frac{\partial \Lcal}{\mu_k} ^\top \cdot \frac{\partial \mu_k}{\partial \alpha_{u, v}}.
	\label{eq:alpha_grad}
\end{equation}
In order to attack the model, we could perform the gradient ascent, \ie, $\alpha_{u, v} \leftarrow \alpha_{u, v} + \eta  \frac{\partial \Lcal}{\partial \alpha_{u, v}}$. However, the attack is on a discrete structure, where only $m$ edges are allowed to be added or deleted. So here we need to solve a combinatorial optimization problem: 
\begin{eqnarray}
	\max_{\{u_t, v_t\}_{t=1}^m} && \sum_{t=1}^m |\frac{\partial \Lcal}{\partial \alpha_{u_t, v_t}}| \nonumber \\
	s.t. &&  \tilde{G} = \text{Modify}(G, \{\alpha_{u_t, v_t}\}_{t=1}^m)\nonumber \\
	&& \Ical(G, \tilde{G}, c) = 1 .
	\label{eq:attack_obj}
\end{eqnarray}
We simply use a greedy algorithm to solve the above optimization. 
Here the modification of $G$ given a set of coefficients $\{\alpha_{u_t, v_t}\}_{t=1}^m$ is performed by sequentially modifying edges $(u_t, v_t)$ of graph $\hat{G}_t$:
\begin{equation}
	\hat{G}_{t+1} = \begin{cases}
		(\hat{V}_t, \hat{E}_t \setminus (u_t, v_t)): \frac{\partial \Lcal}{\partial \alpha_{u_t, v_t}} < 0 \\ 
		(\hat{V}_t, \hat{E}_t \cup \{(u_t, v_t)\}): \frac{\partial \Lcal}{\partial \alpha_{u_t, v_t}}  > 0 \\
	\end{cases}
\end{equation}
That is to say, we modify the edges who are most likely to cause the change to the objective. Depending on the sign of the gradient, we either add or delete the edge. We name it as \textit{GradArgmax} since it does the greedy selection based on gradient information. 

The attack procedure is shown in Figure~\ref{fig:grad_attack}. 
Since this approach requires the gradient information, we consider it as a white-box attack method. Also, the gradient considers all pairs of nodes in a graph, the computation cost is at least $O(|V|^2)$, excluding the back-propagation of gradients in Eq~(\ref{eq:alpha_grad}). Without further approximation, this approach cannot scale to large graphs. 

\subsubsection{Genetic algorithm}
\label{sec:genetic_alg}

\begin{figure}[t]
\centering
\includegraphics[width=0.43\textwidth]{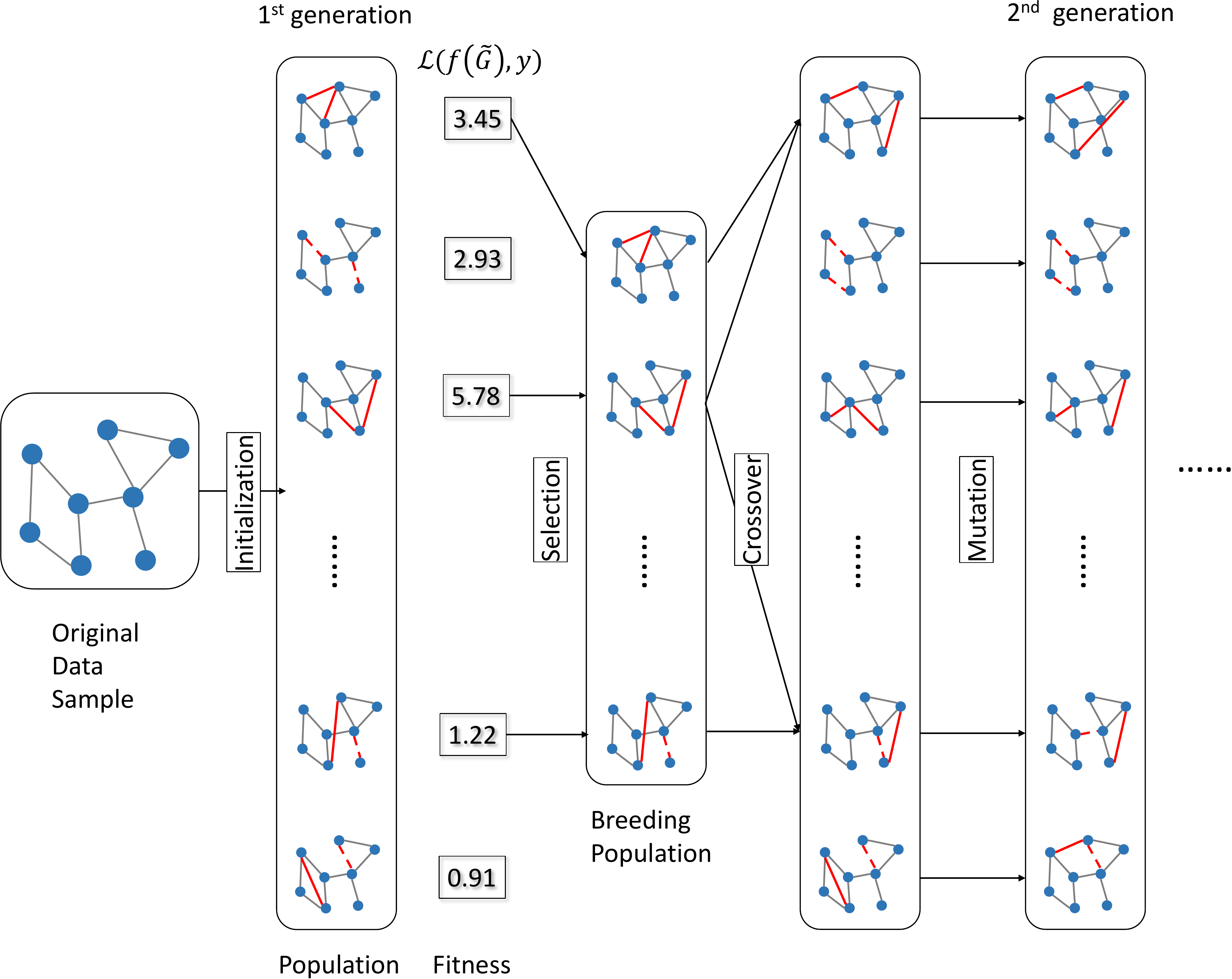}
\vspace{-4mm}
\caption{Illustration of attack using genetic algorithm. The population evolves with selection, crossover and mutation operations. Fitness is measured by the loss function. \label{fig:genetic}}
\end{figure}

Evolution computing has been successfully applied in many zero-order optimization scenarios, including neural architecture search~\citep{ReaMooSelSaxetal17, MiiLiaMeyRawetal17} and adversarial attack for images~\citep{SuVarKou17}. We here propose a black-box attack method that implements a type of genetic algorithms. Given an instance $(G, c, y)$ and the target classifier $f$, Such algorithm involves five major components, as elaborated below:
\begin{itemize}[leftmargin=*,nosep,nolistsep]
\item \textbf{Population}: the population refers to a set of candidate solutions. Here we denote it as $\Pcal^{(r)} = \{\hat{G}_j^{(r)}\}_{j=1}^{|\Pcal^{(r)}|}$, where each $\hat{G}_j^{(r)}$ is a valid modification solution to the original graph $G$. $r=1, 2, \ldots, R$ is the index of generation and $R$ is the maximum numbers of evolutions allowed. 
\item \textbf{Fitness}: each candidate solution in current population will get a score that measures the quality of the solution. We use the loss function of target model $\Lcal(f(\hat{G}_j^{(r)}, c), y)$ as the score function. A good attack solution should increase such loss. Since the fitness is a continuous score, it is not applicable in PBA-D setting, where only classification label is accessible. 
\item \textbf{Selection}:  Given the fitness scores of current population, we can either do weighted sampling or greedy selection to select the `breeding' population $\Pcal^{(r)}_b$ for next generation. 
\item \textbf{Crossover}: After the selection of $\Pcal^{(r)}_b$, we randomly pick two candidates $\hat{G}_1, \hat{G}_2 \in \Pcal^{(r)}_b$ and do the crossover by mixing the edges from these two candidates:
	\begin{equation}
		\hat{G}' = (V, (\hat{E}_1 \cap \hat{E}_2) \cup \text{rp}(\hat{E}_1 \setminus \hat{E}_2) \cup \text{rp}(\hat{E}_2 \setminus \hat{E}_1) ).
	\end{equation}
	Here rp$(\cdot)$ means randomly picking a subset. 
\item \textbf{Mutation}: the mutation process is also biology inspired. For a candidate solution $\hat{G} \in \Pcal^{(r)}$, suppose the modified edges are $\delta E = \{(u_t, v_t)\}_{t=1}^m$. Then for each   edge $(u_t, v_t)$, we have a certain probability to change it to either $(u_t, v')$ or $(u', v_t)$. 
\end{itemize}

The population size $|\Pcal^{(r)}|$, the probability of crossover used in $\text{rp}(\cdot)$, the mutation probability and the number of evolutions $R$ are all hyper-parameters that can be tuned. Due to the limitation of the fitness function, this method can only be used in the PBA-C setting. Also since we need to execute the target model $f$ to get fitness scores, the computation cost of such genetic algorithm is $O(|V| + |E|)$, which is mainly made up by the computation cost of GNNs. The overall procedure is illustrated in Figure~\ref{fig:genetic}. We simply name it as \textit{GeneticAlg} since it is an instantiation of general genetic algorithm framework. 

\begin{table} [t]
\small
\centering
\caption{Application scenarios for different proposed graph attack methods. Cost is measured by the time complexity for proposing a single attack. \label{tab:checklist}}
\vspace{1mm}
\resizebox{0.48\textwidth}{!}{%
\begin{tabular}{c c c c c c c}
 \toprule
 & WBA & PBA-C & PBA-D & RBA & Cost & \\ \midrule
 RandSampling & & & $\surd$ & $\surd$ & $O(1)$ \\
 GradArgmax  &$\surd$ &  & &  & $O(|V|^2)$  \\
 GeneticAlg &  & $\surd$ &  & & $O(|V| + |E|)$ \\
 \textit{RL-S2V} &  & $\surd$  & $\surd$   & $\surd$  & $O(|V|+|E|)$ \\\bottomrule
\end{tabular}
}
\vspace{-3mm}
\end{table}

\vspace{-2mm}
\section{Experiment}
\label{sec:experiment}
\vspace{-2mm}

\begin{table*}[t]
\caption{ Attack graph classification algorithm. We report the 3-class classification accuracy of target model on the vanilla test set I and II, as well as adversarial samples generated. The upper half of the table reports the attack results on test set I, with different levels of access to the information of target classifier. The lower half reports the results of RBA setting on test set II where only \textit{RandSampling} and \textit{RL-S2V} can be used. $K$ is the number of propagation steps used in GNN family models (see Eq~(\ref{eq:gnn})).  \label{tab:component_attack}}
\vspace{-3mm}
\begin{center}
\begin{small}
\resizebox{1.0\textwidth}{!}{%
\begin{tabular}{@{}clcccccccccccc@{}}
\toprule
\multicolumn{2}{c}{attack test set I} &  \multicolumn{4}{c}{15-20 nodes} & \multicolumn{4}{c}{40-50 nodes} & \multicolumn{4}{c}{90-100 nodes} \\ 
\cmidrule(lr){3-6} \cmidrule(lr){7-10} \cmidrule(lr){11-14}
Settings & Methods & $K=2$ & $K=3$ & $K=4$ & $K=5$ & $K=2$ & $K=3$ & $K=4$ & $K=5$ & $K=2$ & $K=3$ & $K=4$ & $K=5$ \\
\midrule
\diagbox{}{} & (unattacked) & 93.20\% &	 98.20\% & 98.87\% &	 99.07\% & 92.60\% & 96.20\% & 97.53\% & 97.93\% & 94.60\% & 97.47\% & 98.73\% & 98.20\% \\
\midrule
RBA & \textit{RandSampling} & 78.73\% & 92.27\% & 95.13\% & 97.67\% & 73.60\% & 78.60\% & 82.80\% & 85.73\% & 74.47\% & 74.13\% & 80.93\% & 82.80\% \\
\midrule
WBA & \textit{GradArgmax} & 69.47\% & 64.60\% & 95.80\% & 97.67\% & 73.93\%& 64.80\% & 70.53\% & 75.47\% & 72.00\% & 66.20\%	 & 67.80\% & 	68.07\%\\
\midrule
PBA-C & \textit{GeneticAlg} & 39.87\% & 39.07\% & 65.33\% & 85.87\% & 59.53\% & 55.67\% & 53.70\% & 42.48\% & 65.47\%	& 63.20\% & 61.60\%	& 61.13\%\\
\midrule
PBA-D & \textit{RL-S2V} & 42.93\% & 41.93\% & 70.20\% & 91.27\% & 61.00\% & 59.20\% & 58.73\% & 49.47\% & 66.07\% & 64.07\% & 64.47\% & 64.67\% \\
\bottomrule
\toprule
\multicolumn{14}{c}{Restricted black-box attack on test set II} \\ 
\midrule
\diagbox{}{} & (unattacked) & 94.67\% & 97.33\% & 98.67\% & 97.33\% & 94.67\% & 97.33\% & 98.67\% & 98.67\% & 96.67\% & 98.00\% & 99.33\% & 98.00\% \\
\midrule
RBA & \textit{RandSampling} & 78.00\% & 91.33\% & 94.00\% & 98.67\% & 75.33\% & 84.00\% & 86.00\% & 87.33\% & 69.33\% & 73.33\% & 76.00\% & 80.00\% \\
\midrule
RBA & \textit{RL-S2V} & 44.00\% & 40.00\% & 67.33\% & 92.00\% & 58.67\% & 60.00\% & 58.00\% & 44.67\% & 62.67\% & 62.00\% & 62.67\% & 61.33\% \\
\bottomrule
\end{tabular}
}
\end{small}
\end{center}
\vspace{-6mm}
\end{table*}

For \textit{GeneticAlg}, we set the population size $|\Pcal|=100$ and the number of rounds $R=10$. We tune the crossover rate and mutation rate in $\{0.1, \ldots, 0.5\}$. For \textit{RL-S2V}, we tune the number of propagations of its S2V model $K=\{1, \ldots, 5\}$. There is no parameter tuning for \textit{GradArgmax} and \textit{RandSampling}.  

We use the proposed attack methods to attack the graph classification model
in Sec~\ref{sec:exp_graph_attack} and node classification model in Sec~\ref{sec:exp_node_attack}. In each scenario, we first show the attack rate when queries are allowed for target model, then we show the generalization ability of the \textit{RL-S2V} for RBA setting.  

\vspace{-2mm}
\subsection{Graph-level attack}
\label{sec:exp_graph_attack}

\begin{figure}[t]
\centering
\begin{tabular}{ccc}
  \includegraphics[width=0.14\textwidth]{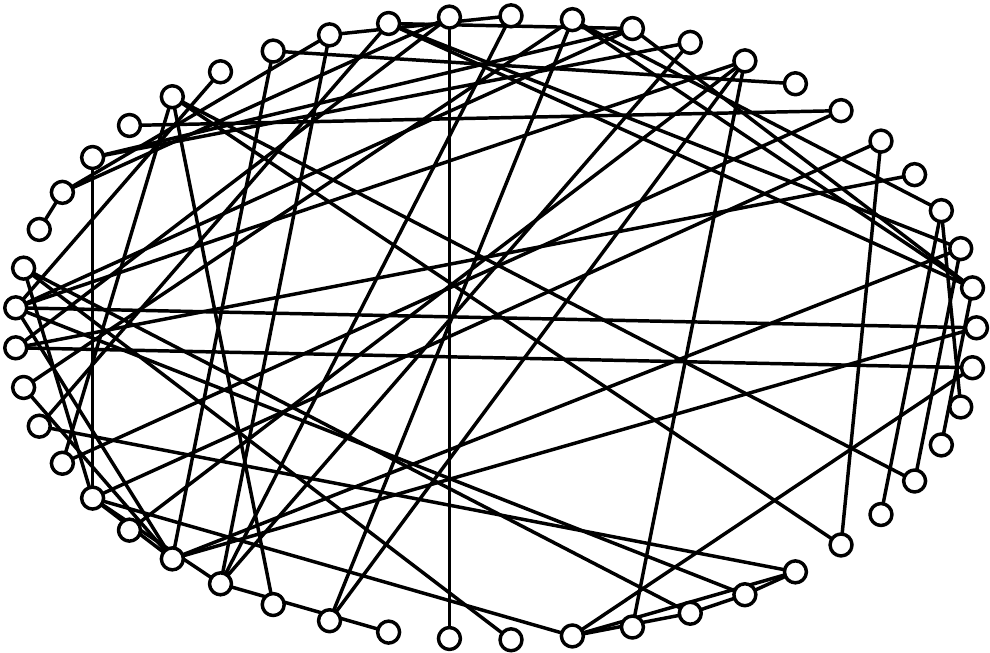} & 
  \includegraphics[width=0.14\textwidth]{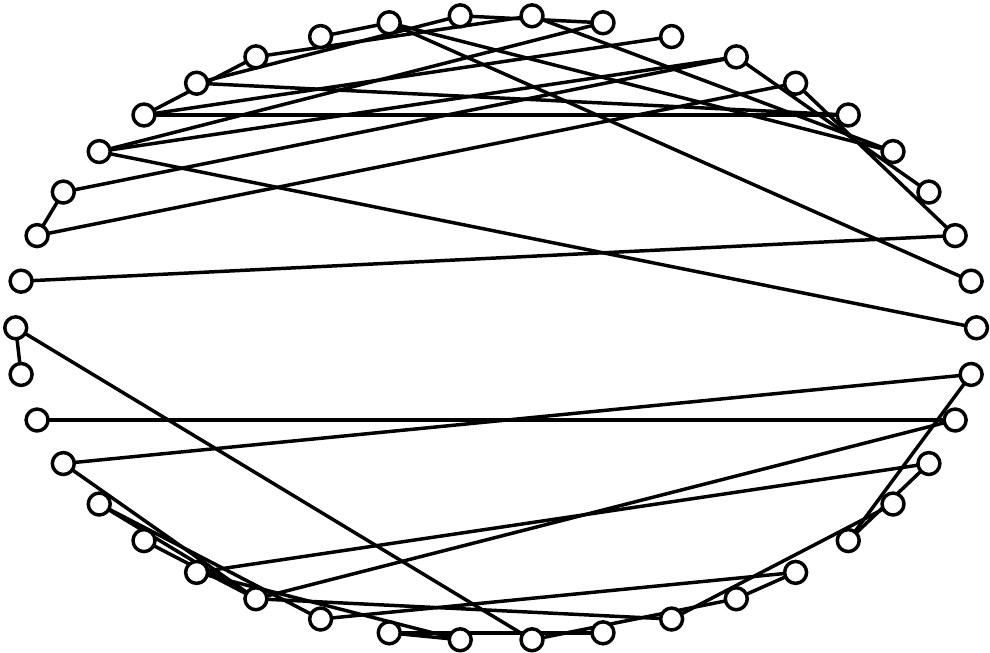} & 
  \includegraphics[width=0.14\textwidth]{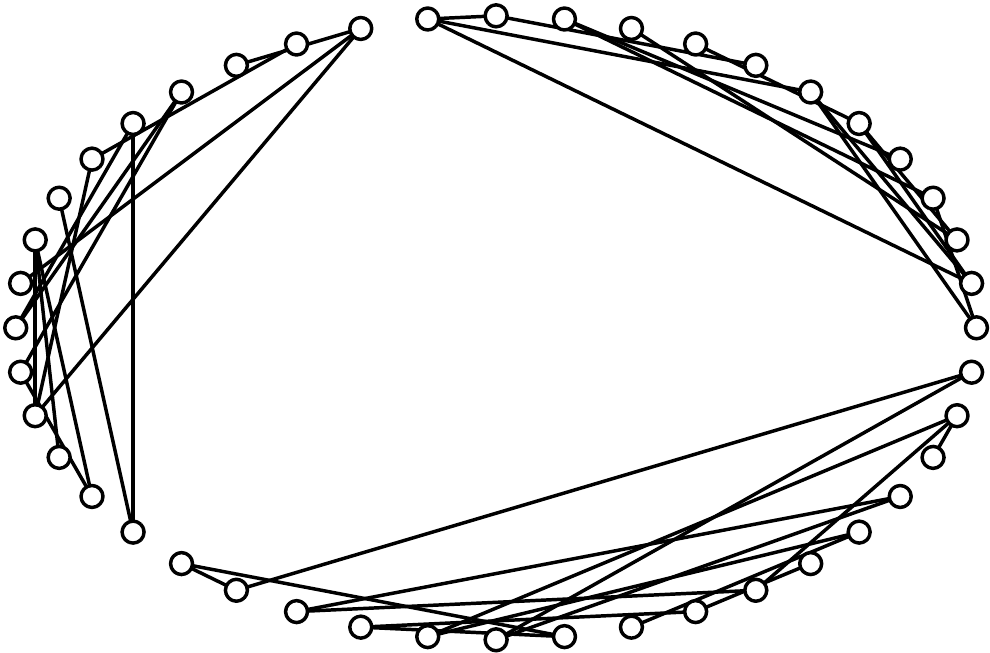} \\
  (a) \# comp $=1$  & (b) \# comp $=2$  & (c) \# comp $=3$ 
\end{tabular}
\vspace{-2mm}
\caption{Example graphs for classification. Here we show three graphs with 1, 2, or 3 components, with 40-50 nodes.   \label{fig:example_component}}
\vspace{-3mm}
\end{figure}

In this set of experiments, we use synthetic data, where the gold classifier $f^*$ is known. Thus the explicit semantics is used for the equivalency indicator $\Ical$. 
The dataset $\Dcal^{(ind)}$ we constructed contains 15,000 graphs, generated with Erdos-Renyi random graph model. It is a three class graph classification task, where each class contains 5,000 graphs. The classifier is asked to tell how many connected components are there in the corresponding undirected graph $G$. The label set $\Ycal = \{1, 2, 3\}$. So there could be up to 3 components in a graph. See Figure~\ref{fig:example_component} for illustration. The gold classifier $f^*$ is obtained by performing a one-time traversal of the entire graph. The dataset is divided into training and two test sets. 
The test set I contains 1,500 graphs, while test set II contains 150 graphs. Each set contains the same number of instances from different classes.

We choose structure2vec as the target model for attack. We also tune its number of propagation parameter $K = \{2, \ldots, 5\}$. Table~\ref{tab:component_attack} shows the results with different settings. For test set I, we can see the structure2vec achieves very high accuracy on distinguishing the number of connected components. Also increasing $K$ seems to improve the generalization in most cases. However, we can see under the practical black-box attack scenario, the \textit{GeneticAlg} and \textit{RL-S2V} can bring down the accuracy to $40\%\sim 60\%$. In attacking the graph classification algorithm, the \textit{GradArgmax} seems not to be very effective. One reason could be the last pooling step in S2V when obtaining graph-level embedding. During back propagation, the pooling operation will dispatch the gradient to every other node embeddings, which makes the $\frac{\partial \Lcal}{\partial \alpha}$ looks similar in most entries. 

For restrict black-box attack on test set II (see the lower half of Table~\ref{tab:component_attack}), the attacker is asked to propose adversarial samples without any access to the target model. Since \textit{RL-S2V} is learned on test set I, it is able to transfer its learned policy to test set II. This suggests that the target classifier makes some form of consistent mistakes. 

This experiment shows that, (1) the adversarial examples do exist for supervised graph problems; (2) a model with good generalization ability can still suffer from adversarial attacks; (3) \textit{RL-S2V} can learn the transferrable adversarial policy to attack unseen graphs.
 
\vspace{-2mm}
\subsection{Node-level attack}
\label{sec:exp_node_attack}
\vspace{-2mm}

In this experiment, we want to inspect the adversarial attack to the node classification problems. Different from Sec~\ref{sec:exp_graph_attack}, here the setting is transductive, where the test samples (but not their labels) are also seen during training. Here we use four real-world datasets, namely the Citeseer, Cora, Pubmed and Finance. The first three are small-scaled citation networks commonly used for node classification, where each node is a paper with corresponding bag-of-words features. The last one is a large-scale dataset that contains transactions from an e-commerce within one day, where the node set contains buyers, sellers and credit cards. The classifier is asked to distinguish the normal transactions from abnormal ones. The statistics of each dataset is shown in Table~\ref{tab:stats}. The nodes also contain features with different dimensions. For the full table please refer to~\citet{KipWel16}. We use GCN~\citep{KipWel16} as the target model to attack. Here the ``small modifications'' is used to regulate the attacker. That is to say, given a graph $G$ and target node $c$, the adversarial samples are limited to delete single edge within 2-hops of node $c$.   

\begin{table}[t]
\centering
\small
\caption{Statistics of the graphs used for node classification. \label{tab:stats}}
\resizebox{0.48\textwidth}{!}{%
\begin{tabular}{cccccc}
\toprule
Dataset & Nodes & Edges & Classes & Train/Test I/Test II \\
\midrule 
Citeseer & 3,327 & 4,732 & 6 & 120/1,000/500 \\
Cora & 2,708 & 5,429 & 7 & 140/1,000/500 \\
Pubmed & 19,717 & 44,338 & 3 & 60/1,000/500 \\
Finance & 2,382,980 & 8,101,757 & 2 & 317,041/812/800 \\
\bottomrule
\end{tabular}
}
\vspace{-3mm}
\end{table}

Table~\ref{tab:attack_node} shows the results. We can see although deleting a single edge is the minimum modification one can do to the graph, the attack rate is still about 10\% on those small graphs, and 4\% in the Finance dataset. We also ran an exhaustive attack as sanity check, which is the best any algorithm can do under the attack budget.
The classifier accuracy will reduce to 60\% or lower if two-edge modification is allowed. However, consider the average degree in the graph is not large, deleting two or more edges would violate the ``small modification'' constraints. We need to be careful to only create adversarial samples, instead of actually changing the true label of that sample. 

In this case, the \textit{GradArgmax} performs quite good, which is different from the case in graph-level attack. Here the gradient with respect to the adjacency matrix $\alpha$ is no longer averaged, which makes it easier to distinguish the useful modifications. 
For the restrict black-box attack on test set II, the \textit{RL-S2V} still learns an attack policy that generalizes to unseen samples. Though we do not have gold classifier in real-world datasets, it is highly possible that the adversarial samples proposed are valid: (1) the structure modification is tiny and within 2-hop; (2) we did not modify the node features. 

\begin{table}[t]
\centering
\small
\caption{Attack node classification algorithm. In the upper half of the table, we report target model accuracy before/after the attack on the test set I, with various settings and methods. In the lower half, we report accuracy on test set II with RBA setting only. In this second part, only \textit{RandSampling} and \textit{RL-S2V} can be used.  \label{tab:attack_node}}
\resizebox{0.49\textwidth}{!}{%
\begin{tabular}{@{}lcccc@{}}
\toprule
   Method & Citeseer & Cora & Pubmed & Finance  \\
\midrule
 (unattacked) & 71.60\% & 81.00\% & 79.90\% & 88.67\% \\
\midrule
RBA, \textit{RandSampling} & 67.60\%	 & 78.50\% & 79.00\% & 87.44\% \\
\midrule
WBA, \textit{GradArgmax} & 63.00\% & 71.30\% & 72.4\% & 86.33\% \\
\midrule
PBA-C, \textit{GeneticAlg} & 63.70\% & 71.20\% & 72.30\% & 85.96\%\\
\midrule
PBA-D, \textit{RL-S2V} & 62.70\% & 71.20\% & 72.80\% & 85.43\%  \\
\midrule
Exhaust & 62.50\% & 70.70\% & 71.80\% & 85.22\%\\
\bottomrule 
\toprule 
\multicolumn{5}{c}{Restricted black-box attack on test set II} \\
\midrule
(unattacked) & 72.60\% & 80.20\% & 80.40\% & 91.88\% \\
\midrule
\textit{RandSampling} & 68.00\% & 78.40\% & 79.00\% & 90.75\% \\
\midrule
\textit{RL-S2V} & 66.00\% & 75.00\% & 74.00\% & 89.10\% \\
\midrule
Exhaust & 62.60\% & 70.80\% & 71.00\% & 88.88\%\\
\bottomrule
\end{tabular}
}
\vspace{-4mm}
\end{table}

\vspace{-2mm}
\subsection{Inspection of adversarial samples}
\vspace{-2mm}

\begin{figure}[t]
\centering
\begin{tabular}{ccc}
  \includegraphics[width=0.14\textwidth]{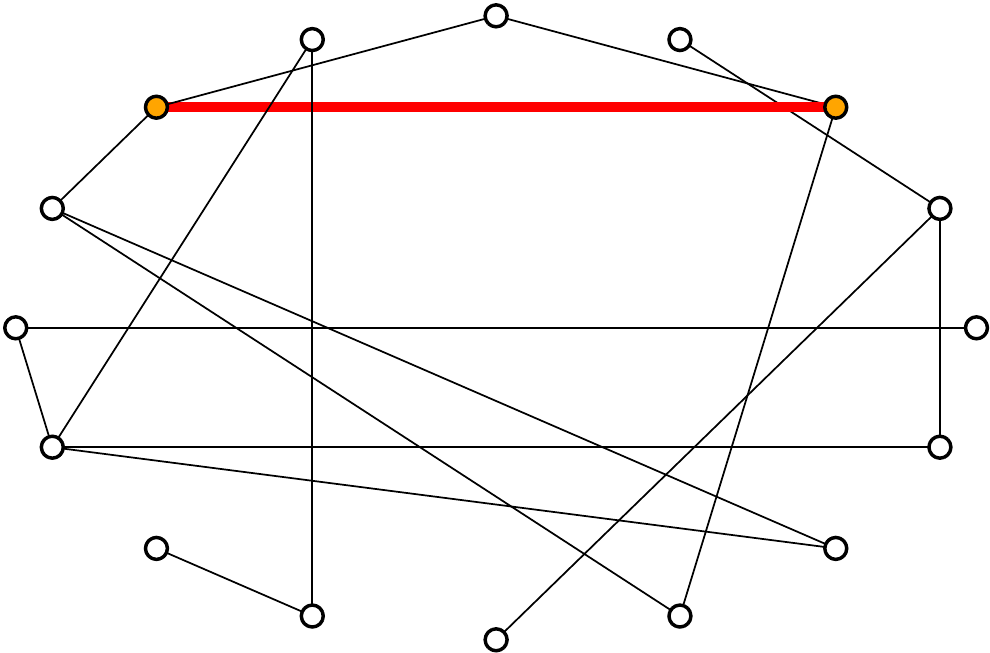} & 
  \includegraphics[width=0.14\textwidth]{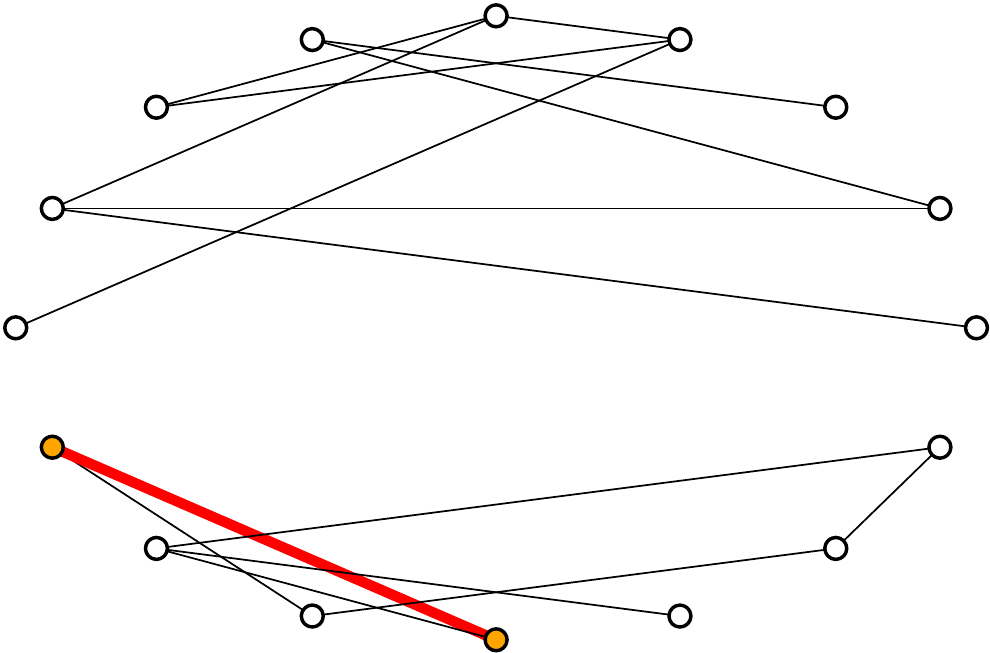} & 
  \includegraphics[width=0.14\textwidth]{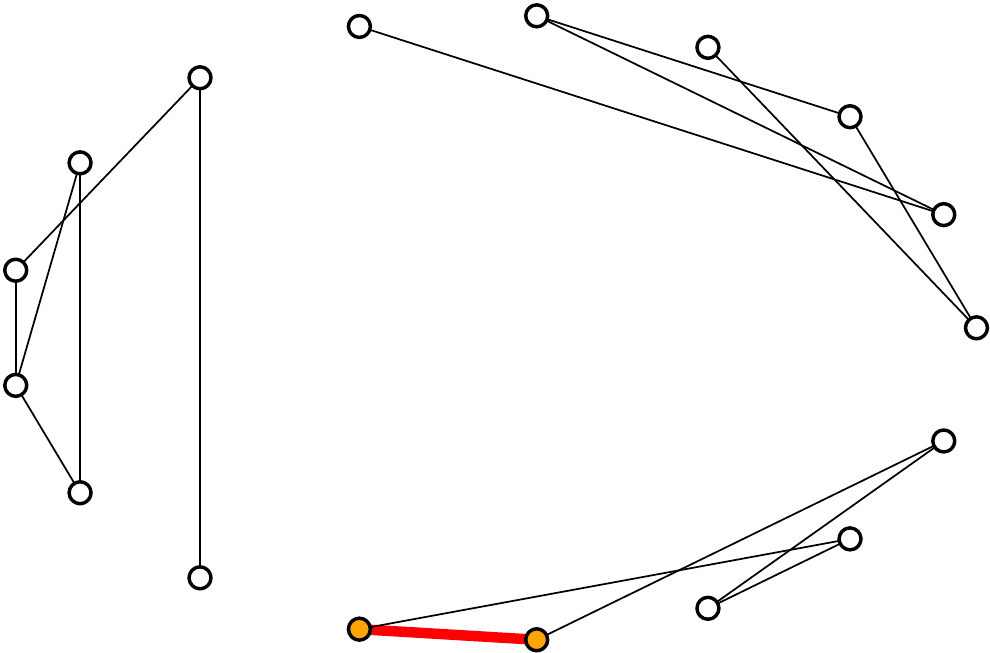} \\
  (a) pred $=2$  & (b) pred $=1$  & (c) pred $=2$ 
\end{tabular}
\vspace{-3mm}
\caption{Attack solutions proposed by \textit{RL-S2V} on graph classification problem. Target classifier is structure2vec with $K=4$. The ground truth \# components are: (a) 1 (b) 2 (c) 3. \label{fig:dqn_demo}}
\vspace{-3mm}
\end{figure}

In this section, we visualize the adversarial samples proposed by different attackers. The solutions proposed by \textit{RL-S2V} for graph-level classification problem are shown in Figure~\ref{fig:dqn_demo}. The ground truth labels are 1, 2, 3, while the target classifier mistakenly predicts 2, 1, 2, respectively. In Figure~\ref{fig:dqn_demo}(b) and (c), the RL agent connects two nodes who are 4 hops away from each other (before the red edge is added). This shows that, although the target classifier structure2vec is trained with $K=4$, it didn't capture the 4-hop information efficiently. Also Figure~\ref{fig:dqn_demo}(a) shows that, even connecting nodes who are just 2-hop away, the classifier makes mistake on it. 

\begin{figure}[t]
\centering
\begin{tabular}{ccc}
  \includegraphics[width=0.14\textwidth]{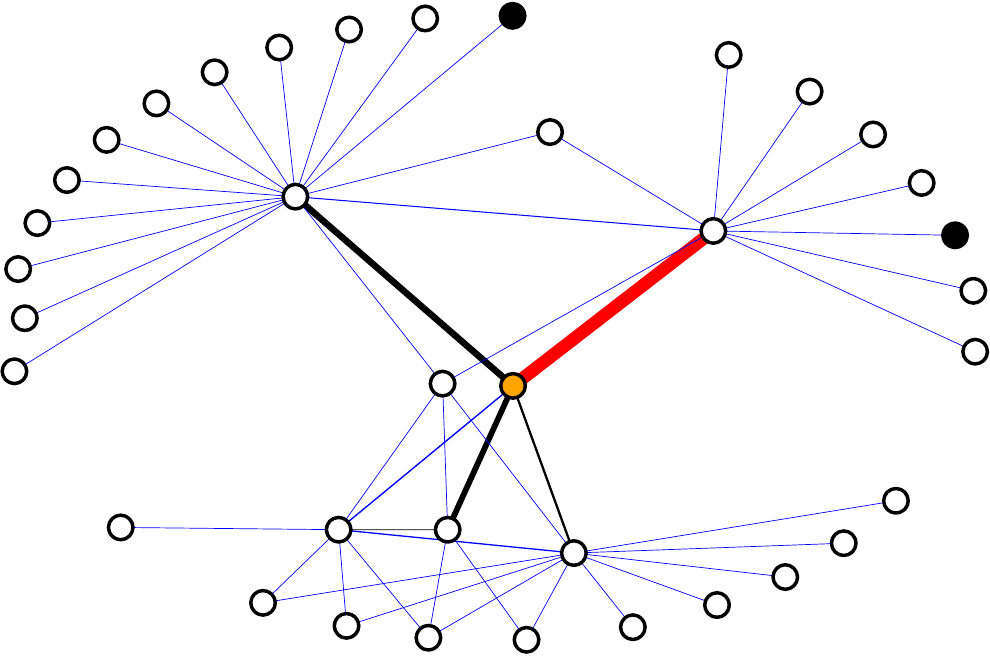} & 
  \includegraphics[width=0.14\textwidth]{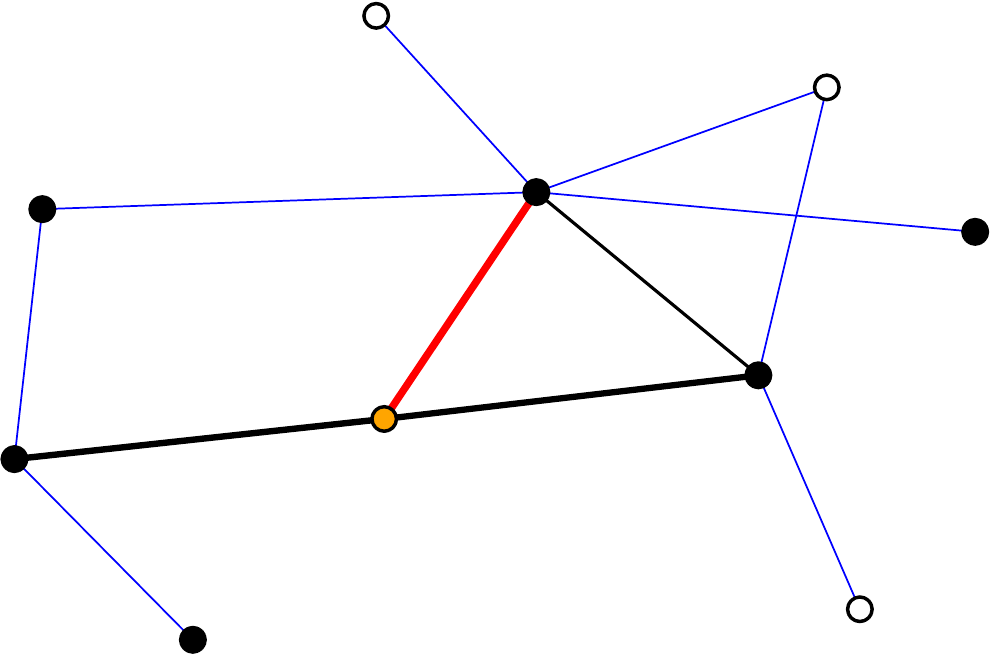} & 
  \includegraphics[width=0.14\textwidth]{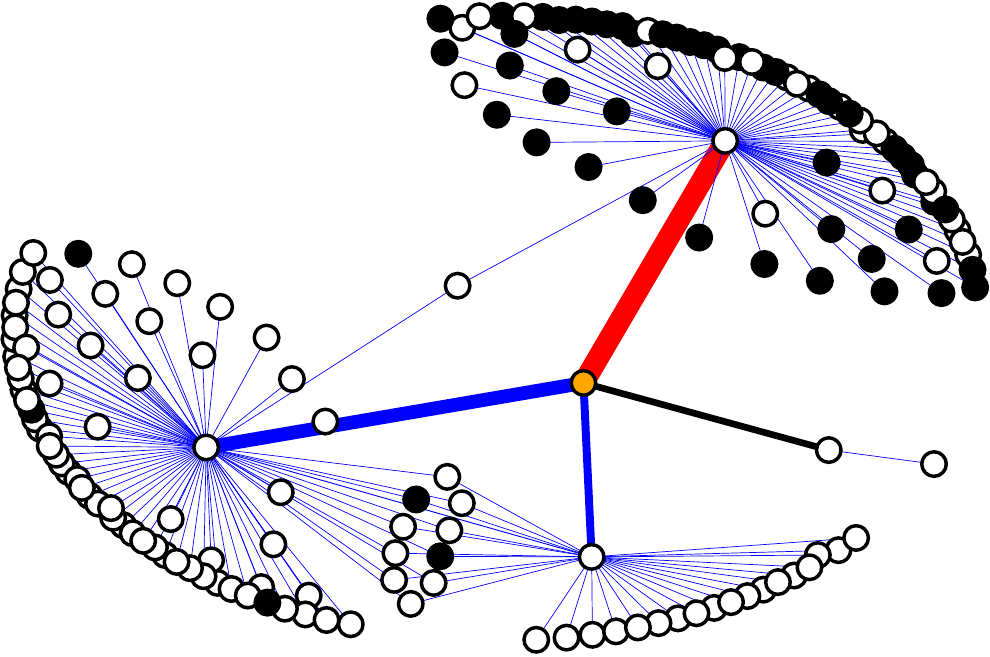} \\
  (a) pred $=2$  & (b) pred $=1$  & (c) pred $=2$ 
\end{tabular}
\vspace{-2mm}
\caption{Attack solutions proposed by \textit{GradArgmax} on node classification problem. Attacked node is colored orange. Nodes from the same class as the attacked node are marked black, otherwise white.  Target classifier is GCN with $K=2$. \label{fig:grad_demo}}
\vspace{-2mm}
\end{figure}

Figure~\ref{fig:grad_demo} shows the solutions proposed by \textit{GradArgmax}. Orange node is the target node for attack. Edges with blue color are suggested to be added by \textit{GradArgmax}, while black ones are suggested to be deleted. Black nodes have the same node label as the orange node, while while nodes do not. The thicker the edge, the larger the magnitude of the gradient is. Figure~\ref{fig:grad_demo}(b) deletes one neighbor with the same label, but still have other black nodes connected. In this case, the GCN is over-sensitive. The mistake made in Figure~\ref{fig:grad_demo}(c) is reasonable, since although the red edge does not connect two nodes with the same label, it connects to a large community of nodes from the same class in 2-hop distance. In this case, the prediction made by GCN is reasonable. 

\vspace{-2mm}
\subsection{Defense against attacks}
\vspace{-2mm}

\begin{table}[t]
\centering
\small
\caption{Results after adversarial training by random edge drop.  \label{tab:adv_train}}
\resizebox{0.48\textwidth}{!}{%
\begin{tabular}{@{}lcccc@{}}
\toprule
   Method & Citeseer & Cora & Pubmed & Finance  \\
\midrule
 (unattacked) & 71.30\% & 81.70\% & 79.50\% & 88.55\% \\
\midrule
RBA, \textit{RandSampling} & 67.70\%	 & 79.20\% & 78.20\% & 87.44\%\\
\midrule
WBA, \textit{GradArgmax} & 63.90\% & 72.50\% & 72.40\% & 87.32\%\\
\midrule
PBA-C, \textit{GeneticAlg} & 64.60\% & 72.60\% & 72.50\% & 86.45\%\\
\midrule
PBA-D, \textit{RL-S2V} & 63.90\% & 72.80\% & 72.90\% & 85.80\% \\
\bottomrule 
\end{tabular}
}
\vspace{-3mm}
\end{table}

Different from the images, here the possible number of graph structures is finite given the number of nodes. So by adding the adversarial samples back for further training, the improvement of the target model's robustness can be expected. For example, in the experiment of Sec~\ref{sec:exp_graph_attack}, adding adversarial samples for training is equivalent to increasing the size of the training set, which will definitely be helpful. So here we seek to use a cheap method for adversarial training --- simply doing edge drop during training for defense. 

Dropping the edges during training is different from Dropout~\citep{SriHinKriSutSal14}. Dropout operates on the neurons in the hidden layers, while edge drop modifies the discrete structure. It is also different from simply drop the entire hidden vector, since deleting a single edge can affect more than just one edge. For example, GCN computes the normalized graph Laplacian. So after deleting a single edge, the normalized graph Laplacian needs to be recomputed for some entries. 
This approach is similar to~\citet{HamYinLes17}, who samples a fixed number of neighborhoods during training for the efficiency. Here we drop the edges globally at random, during each training step. 

The new results after adversarial training are presented in Table~\ref{tab:adv_train}. We can see from the table that, though the accuracy of target model remains similar, the attack rate of various methods decreases about $1\%$. Though the scale of the improvement is not significant, it shows some effectiveness with such cheap adversarial training. 

\vspace{-2mm}
\section{Related work}
\vspace{-2mm}

\noindent \textbf{adversarial attack in continuous and discrete space}: 
In recent years, the adversarial attacks to the deep learning models have raised increasing attention from researchers. Some methods focus on the white-box adversarial attack using gradient information, like box constrained L-BFGS~\citep{SzeZarSutBruetal13}, 
Fast Gradient Sign~\citep{GooShlSze14}, deep fool~\citep{MooFawFro16}, \etc. When the full information of target model is not accessible, one can train a substitute model\citep{PapMcdGooJha17}, or use zero-order optimization method~\citep{CheZhaShaYietal17}. 
There are also some works working on the attack with discrete functions~\citep{buckman2018thermometer} but not the combinatorial structures. The one-pixel attack~\citep{SuVarKou17} modifies the image by only several pixels using differential evolution; ~\citet{JiaLiang17} attacks the text reading comprehension system with the help of rules and human efforts. ~\citet{zugner2018adversarial} studied the problem of adversarial attack over graphs in parallel to our work, although with very different methods. \\
\noindent \textbf{combinatorial optimization}: 
Modifying the discrete structure to fool the target classifier can be treated as a combinatorial optimization problem. Recently, there are some exciting works using reinforcement learning to learn to solve the general sequential decision problems~\citep{BelPhaLeNoretal16} or graph combinatorial problems~\citep{DaiKhaZhaDiletal17}. These are closely related to \textit{RL-S2V}. The \textit{RL-S2V} extends the previous approach using hierarchical way to decompose the quadratic action space, in order to make the training feasible. 

\vspace{-3mm}
\section{Conclusion}
\vspace{-2mm}

In this paper, we study the adversarial attack on graph structured data. To perform the efficient attack, we proposed three methods, namely \textit{RL-S2V}, \textit{GradArgmax} and \textit{GeneticAlg} for three different attack settings, respectively. 
We show that a family of GNN models are vulnerable to such attack. By visualizing the attack samples, we can also inspect the target classifier. We also discussed about defense methods through experiments. Our future work includes developing more effective defense algorithms.



%

\newpage
\section*{Acknowledgements}

This project was supported in part by NSF IIS-1218749, NIH BIGDATA 1R01GM108341, NSF CAREER IIS-1350983, NSF IIS-1639792 EAGER, NSF CNS-1704701, ONR N00014-15-1-2340, Intel ISTC, NVIDIA and Amazon AWS. Tian Tian and Jun Zhu were supported by the National NSF of China (No. 61620106010) and Beijing Natural Science Foundation (No. L172037). We thank Bo Dai for valuable suggestions, and the anonymous reviewers who gave useful comments.
%
%



\bibliographystyle{icml2018}




\end{document}